\documentclass{article}
\usepackage{spconf,amsmath,graphicx,subfigure,caption}
\usepackage{booktabs}
\usepackage{enumerate}
\usepackage{enumitem}
\usepackage{hyperref}
\setlist[enumerate]{itemsep=0mm}
\title{Unified learning-based lossy and lossless JPEG recompression\thanks{*Corresponding author.}}
\name{
\em Jianghui Zhang$^{1}$, Yuanyuan Wang$^{2}$, Lina Guo$^{2}$, Jixiang Luo$^{2}$, Tongda Xu$^{1}$, \\
\em Yan Wang$^{1*}$, Zhi Wang$^1$, Hongwei Qin$^2$
}
\address{
$^1$Tsinghua University \\
$^2$SenseTime Research
}
\begin{document}
\maketitle
\begin{abstract}
    JPEG is still the most widely used image compression algorithm. Most image compression algorithms only consider uncompressed original image, while ignoring a large number of already existing JPEG images. Recently, JPEG recompression approaches have been proposed to further reduce the size of JPEG files. However, those methods only consider JPEG lossless recompression, which is just a special case of the rate-distortion theorem. In this paper, we propose a unified lossly and lossless JPEG recompression framework, which consists of learned quantization table and Markovian hierarchical variational autoencoders. Experiments show that our method can achieve arbitrarily low distortion when the bitrate is close to the upper bound, namely the bitrate of the lossless compression model. To the best of our knowledge, this is the first learned method that bridges the gap between lossy and lossless recompression of JPEG images.
\end{abstract}

\section{Introduction}
With the rapid development of the internet and smart devices, multimedia data is growing exponentially. Due to the expansive cost of transmission and storage, data compression is indispensable for data centers, cloud storage, and network filesystems. JPEG~\cite{pennebaker1992jpeg} is a commonly used method of lossy compression for digital images, which was first published in 1992. After decades of technological development, a variety of new image compression algorithms are emerging, like JPEG2000~\cite{rabbani2002jpeg2000}, PNG~\cite{boutell1997png}, WEBP~\cite{lian2012webp}, BPG~\cite{bellard2015bpg}, the intra-coding of VVC/H.266~\cite{ohm2018versatile} and learning based methods~\cite{Balle2017ICLR, Balle2018ICLR, Minnen2018NIPS, Cheng2020CVPR, Gao2021ICCV, he2022elic}. However, these subsequent methods only focus on uncompressed original image data, ignoring the large number of JPEG images that already exist on the internet. According to the survey~\cite{horn2017design} of Dropdox, JPEG files take up about 35\% of the total storage space. 

\begin{figure}
    \centering
    \includegraphics[scale=0.48]{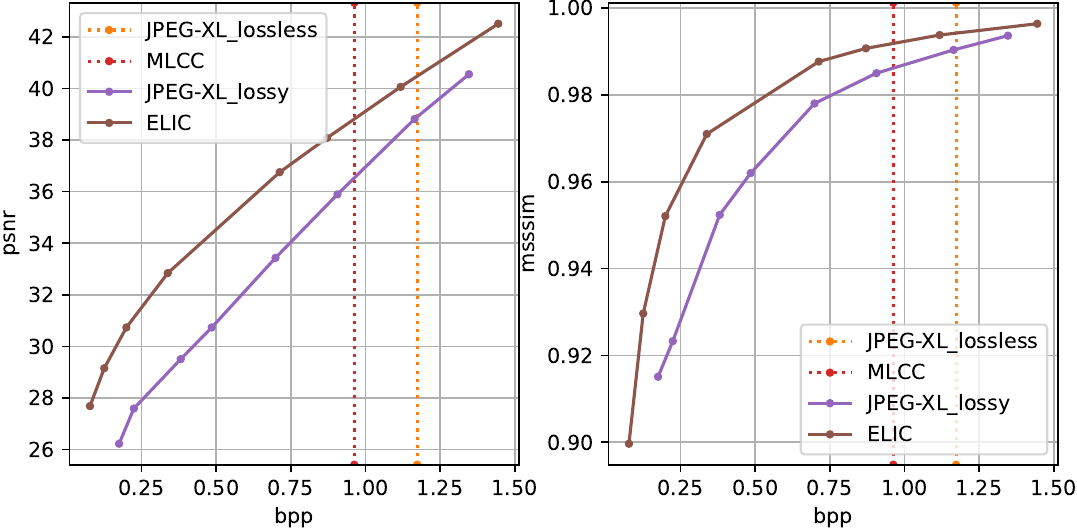}
    \caption{Objective quality comparison on Kodak dataset when recompressing JPEG images of QP 75 with different methods: JPEG-XL~\cite{alakuijala2019jpeg}, MLCC~\cite{guo2022practical} and ELIC~\cite{he2022elic}.}
    \label{fig:Reverse-Compression}
    \vspace{-1em}
\end{figure}

Due to the demand to further reduce the size of JPEG files, several methods for JPEG recompression are proposed, e.g. Lepton~\cite{horn2017design}, JPEG-XL~\cite{alakuijala2019jpeg} 
and MLCC~\cite{guo2022practical}. However, these methods only consider the lossless recompression for JPEG files, which is a special case of the rate-distortion theorem. In other words, those methods cannot provide variable bitrates for different distortion requirements, which means lossy recompression of JPEG files. 

Intuitively, those new lossy compression methods~\cite{lian2012webp, bellard2015bpg, alakuijala2019jpeg, Balle2017ICLR, Balle2018ICLR, Minnen2018NIPS, Cheng2020CVPR, Gao2021ICCV} can also be used for lossy JPEG recompression. Specifically, the JPEG file is firstly decompressed to RGB data, and we can use those new lossy compression methods to compress the RGB data. However, this straightforward method will induce the \textbf{Reverse-Compression} problem when the desired distortion is below a certain value. As shown in Fig. \ref{fig:Reverse-Compression}, whether we use hand-designed compression methods or learnable compression methods to recompress JPEG files in this way, there is still non-negligible compression distortion when the bitrate exceeds the bitrate of lossless recompression, which should be the upper bound for any lossy recompression. According to the rate-distortion theory~\cite{yeung2008information}, it means  there is a gap between current lossless recompression methods and lossy recompression methods for JPEG images.

In this paper, we propose a unified JPEG lossy and lossless recompression framework, which consists of one learned quantization table, one learned inverse quantization table and a DCT-domain lossless compression model implemented by Markovian hierarchical variational autoencoder.
The contributions of this paper are summarized as follows:
\begin{enumerate}
    \item We proposed an end-to-end learned framework for unified lossy and lossless JPEG recompression. To the best of our knowledge, this is the first approach that bridges the gap between lossy recompression and lossless recompression for JPEG images.
    \item Experiments show that our method for lossy recompression can achieve arbitrarily low distortion while keeping the bitrate always below the upper bound, $i.e.$ the bitrate of lossless recompression. 
\end{enumerate}

\section{Related Work}
\vspace{-0.5em}
\subsection{Overview of JFIF standard}
As one of the most common image formats, JFIF~\cite{pennebaker1992jpeg} defines supplementary specifications for the JPEG algorithm. During the compression phase, JFIF first converts RGB signals into YCbCr color space. Then, since the human visual system is more sensitive to the luma component compared to the chroma component, commonly used YCbCr 4:2:0 subsampling is applied to reduce data redundancy. After that, each component is divided into $8 \times 8$ blocks. For each $8 \times 8$ block, discrete cosine transformation (DCT) is applied to get frequency data, so-called dct coefficients. Subsequently, these three components are quantized with two quantization tables. Y component is quantized with one quantization table, while CbCr components share another quantization table. Finally, quantized dct coefficients are compressed with lossless Huffman coding. The decompression process is the reverse of the compression process. 
For the sake of clarity, in the following, we denote the \textbf{QP 75} as the meaning of using the example table defined by Annex K of the JPEG standard~\cite{pennebaker1992jpeg} with scale factor QP=75 according to the IJG scaling algorithm.

\vspace{-0.5em}
\subsection{JPEG recompression methods}
\textbf{Lepton}~\cite{horn2017design} was proposed by Horn \textit{et al.}, which mainly focuses on the optimization of the entropy model and symbol representations. Lepton achieves more than $20\%$ storage saving for JPEG lossless recompression.
\quad\\
\textbf{JPEG-XL}~\cite{alakuijala2019jpeg} is a new image coding standard, which supports both lossless and lossy compression. For JPEG images, lossless recompression is also supported by JPEG-XL. JPEG-XL achieves better compression ratio by a series of optimization, like variable-size DCT block, local adaptive quantization, more DC coefficient prediction modes, and using Asymmetric Numeral Systems in place of Huffman coding.
\quad\\
\textbf{MLCC}~\cite{guo2022practical} proposed by Guo \textit{et al.} is the first learned lossless JPEG recompression method. They proposed a Multi-Level Cross-Channel entropy model for lossless recompression of JPEG images, which achieves state-of-the-art performance. For JPEG images with QP 75, MLCC obtains about $30\%$ compression savings.

\vspace{-0.5em}
\subsection{Learned image compression}
\textbf{Lossy Compression}
In recent years, since Balle \textit{et al.}~\cite{Balle2017ICLR} first proposed an end-to-end image compression model based on autoencoder architecture, a series of subsequent methods~\cite{Balle2018ICLR, Balle2018ICLR, Minnen2018NIPS, Cheng2020CVPR, Gao2021ICCV, he2022elic} are proposed to improve the compression ratio. They mainly focus on the optimization of rate estimation with non-differential quantization and entropy estimation by designing a more accurate entropy model. The newest method ~\cite{he2022elic} has outperformed the intra-coding of VVC/H.266~\cite{ohm2018versatile}, which is the best hand-crafted method.
\quad\\
\textbf{Lossless Compression}
Lossless compression is highly relevant to the probabilistic model. Theoretically, any probabilistic model combined with an entropy encoder can be used for lossless compression. With the development of probabilistic models, lossless compression methods~\cite{reed2017ms-pixelcnn, kingma2019bitswap, mentzer2019l3c, berg2021idf++} have also made great progress. The recent method~\cite{mentzer2019l3c} has outperformed the popular engineered lossless compression codecs, like PNG~\cite{boutell1997png}, WebP~\cite{lian2012webp} and JPEG2000~\cite{rabbani2002jpeg2000}.

However, all of these methods only consider the compression of RGB image data, and cannot directly be applied to JPEG image data ~\cite{guo2022practical}.

\begin{figure}[t]
    \centering
    \includegraphics[scale=0.55]{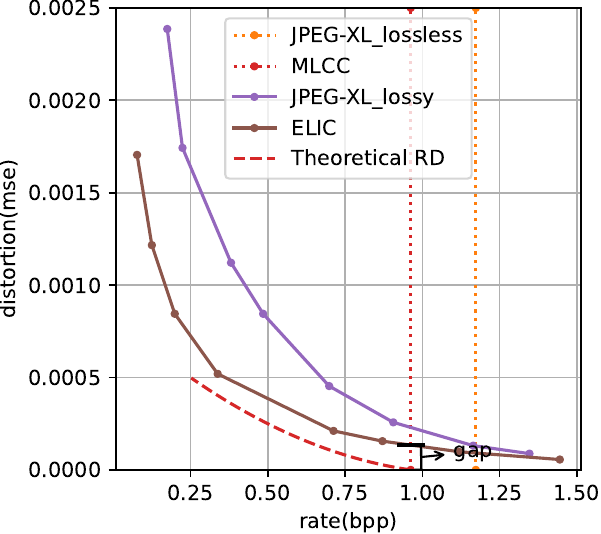}
    \caption{Rate-distortion of lossless compression models and lossy compression models for JPEG QP 75 recompression.}
    \label{fig:rate-distortion}
    \vspace{-1em}
\end{figure}

\section{Method}
\vspace{-0.5em}
\subsection{Motivation}
For the rate-distortion problem of JPEG recompression, without losing generality, we consider bpp as the rate metric and mean squared error as the distortion metric. Assuming the range of each pixel value is $[0, 1]$. Let's take the latest lossy compression method ELIC~\cite{he2022elic} and lossless compression method for JPEG recompression MLCC~\cite{guo2022practical} as an example. For the sake of convenience, we denote the rate of MLCC as $H$. As shown in Fig. \ref{fig:rate-distortion}, there is a remarkable gap between the rate-distortion curve of ELIC and the rate of MLCC. According to rate-distortion theory~\cite{yeung2008information}, there must be one theoretical convex rate-distortion curve (marked as Theoretical RD in Fig. \ref{fig:rate-distortion}) for the MLCC-based lossy compression method, which can achieve arbitrarily low distortion when the rate is close to $H$. In other words, theoretically, there is a better method for JPEG recompression compared to ELIC when the rate is close to $H$. 

%TODO: polish
To sum up, as shown in Fig. \ref{fig:rate-distortion}, neither hand-designed lossy compression methods like JPEG-XL nor learnable lossy compression methods like ELIC performs well for JPEG lossy recompression. Once the rate exceeds $H$, it is better to use MLCC rather than those lossy compression methods for JPEG recompression. 

\begin{figure*}[!htbp]
    \centering
    \includegraphics[scale=0.56]{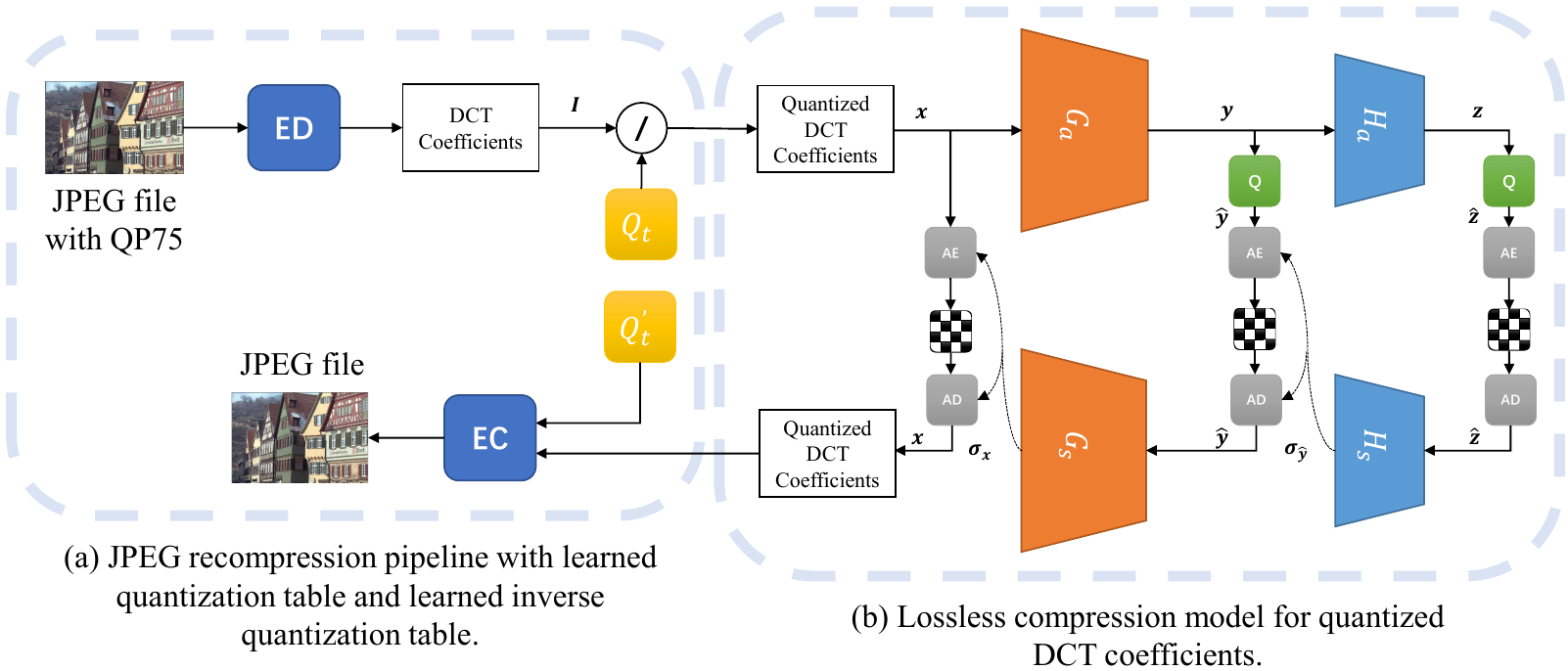}
    \caption{Overall Architecture of our method. \textbf{ED} denotes entropy decoder followd by inverse quantization operation, \textbf{EC} denotes entropy encoder. $Q_t$ and $Q_t^{'}$ represent the learned quantization table and learned inverse quantization table respectively. The lossless compression model for quantized DCT coefficients is adapted from ~\cite{Balle2018ICLR}, where \textbf{Q} denotes quantization, \textbf{AE} and \textbf{AD} denote arithmetic coding and arithmetic decoding respectively. The architecture of $G_a$, $G_s$, $H_a$, $H_s$ is in Tab. \ref{tab:lossless model}.}
    \label{fig:architecture}
    \vspace{-1em}
\end{figure*}

\vspace{-0.5em}
\subsection{Framework}
The framework of our method is shown in Fig.~\ref{fig:architecture}. 
For a given JPEG file with QP 75, it is first decoded to DCT coefficients with the entropy decoder followed by inverse quantization operation. Then, the learned quantization table is applied to quantize the DCT coefficients. Quantized DCT coefficients are losslessly compressed into the bitstream by a lossless compression model with Markovian hierarchical variational autoencoder architecture, which is adapted from ~\cite{Balle2018ICLR}. During the decompression phase, the bitstream is first decompressed to quantized DCT coefficients. And then, it is compressed into a JPEG file together with the learned inverse quantization table by the entropy encoder. 
% Actually, the lossless compression model in our framework can be replaced with any other learned lossless compression for DCT-domain data compression, like MLCC. 
% Due to the complexity to reproduce MLCC, whose author doesn't release their code, we choose to modify the model ~\cite{Balle2018ICLR} as our lossless compression model. 

In our framework, the lossless compression model is used to optimize the rate, the learned quantization table and inverse quantization table are used to optimize the rate and distortion. We adopt iterative optimization way to optimize each of these two items separately. Given a JPEG image as input, we denote its DCT coefficients as $I$, the learned quantization table as $Q_t$, the learned inverse quantization table as $Q_t^{'}$ and the parameters of the lossless model as $\theta$, respectively. Then the input of the lossless compression model as shown in Fig.~\ref{fig:architecture} is $\boldsymbol{x} = \left\lfloor \frac{I}{Q_t} \right\rceil$, $\lfloor \cdot \rceil$ denotes straight-through estimator. And there are two hidden variables $\boldsymbol{y}, \boldsymbol{z}$ in the framework, where $\boldsymbol{y} = G_a(\boldsymbol{x})$ and $\boldsymbol{z} = H_a(\boldsymbol{y})$. The loss function for optimizing the lossless compression model is formulated as follows:
\begin{equation}
    \begin{aligned}
        \mathcal{L}_{m} = R_\theta(\boldsymbol{x}) \\
    \end{aligned}
    \label{eq:L_m}
\end{equation}
where $Q_t$ is fixed during optimizing $\theta$.
The loss function for optimizing the learned quantization table and inverse quantization table is formulated as follows:
\begin{equation}
    \begin{aligned}
        \mathcal{L}_{q} = \lambda_r R_\theta(\boldsymbol{x}) + \lambda_d D(I, \boldsymbol{x} \cdot Q_t^{'}) \\
    \end{aligned}
    \label{eq:L_q}
\end{equation}
where $\lambda_r$ and $\lambda_d$ are hyper-parameters used to control the rate-distortion trade-off. The parameters $\theta$ is fixed during optimizing $Q_t$ and $Q_t^{'}$. Following the setting in ~\cite{Balle2018ICLR}, the expected negative entropy is supervised as the $R_\theta$ term and the MSE measure on RGB domain is used as the $D$ term. Specifically, $R_\theta$ term and $D$ term are formulated as follows, respectively: 
\begin{equation}
    \begin{aligned}
        R_{\theta}(\boldsymbol{x}) = & - \mathbf{E} \left[ \log{p(\boldsymbol{x}, \hat{\boldsymbol{y}}, \hat{\boldsymbol{z}})} \right] \\
            = & - \mathbf{E} \left[ \log{p(\boldsymbol{x} | \hat{\boldsymbol{y}})} + \log{p(\hat{\boldsymbol{y}} | \hat{\boldsymbol{z}})} + \log{p(\hat{\boldsymbol{z}})} \right] \\
        D(I, \boldsymbol{x} \cdot Q_t^{'}) & = \mathbf{E} \left[\left( f_{idct}(I) - f_{idct}(\boldsymbol{x} \cdot Q_t^{'}) \right)^2 \right]
    \end{aligned}
\end{equation}
where $\hat{\boldsymbol{y}} = \lfloor \boldsymbol{y} \rceil, \hat{\boldsymbol{z}} = \lfloor \boldsymbol{z} \rceil$, $f_{idct}$ denotes inverse discrete cosine transformation, $p(\boldsymbol{x} | \hat{\boldsymbol{y}})$, $p(\hat{\boldsymbol{y}} | \hat{\boldsymbol{z}})$ are respectively modeled as $i.i.d.$ gaussian distribution with mean 0, variance $\sigma_{\boldsymbol{x}} = G_s(\hat{\boldsymbol{y}})$, $\sigma_{\hat{\boldsymbol{y}}} = H_s(\hat{\boldsymbol{z}})$ and $p(\hat{\boldsymbol{z}})$ is modeled as a fully factorized density model~\cite{Balle2018ICLR}.

\begin{table}[ht]
    \centering
    \begin{tabular}{cc}
        \hline
        $G_a$ & $G_s$ \\
        \hline
        Conv $5\!\! \times\!\! 5$, s2 $\mid$ ReLU & Conv $3\!\! \times\!\! 3$, s1 $\mid$ ReLU \\
        Conv $5\!\! \times\!\! 5$, s2 $\mid$ ReLU6 & Conv $3\!\! \times\!\! 3$, s1 $\mid$ ReLU \\
        Conv $5\!\! \times\!\! 5$, s2 $\mid$ ReLU6 & Conv $3\!\! \times\!\! 3$, s1 \\
        Conv $5\!\! \times\!\! 5$, s2 & ConvTranspose $4\!\! \times\!\! 4$, s2 $\mid$ ReLU \\
         & ConvTranspose $4\!\! \times\!\! 4$, s2 $\mid$ ReLU \\
         & ConvTranspose $4\!\! \times\!\! 4$, s4 \\
        \hline
    \end{tabular}
    \smallskip \\
    \begin{tabular}{cc}
        \hline
        $H_a$ & $H_s$ \\
        \hline
        Conv $3\!\! \times\!\! 3$, s1 $\mid$ ReLU & ConvTranspose $4\!\! \times\!\! 4$, s2 $\mid$ ReLU  \\
        Conv $5\!\! \times\!\! 5$, s2 $\mid$ ReLU & ConvTranspose $4\!\! \times\!\! 4$, s2 $\mid$ ReLU \\
        Conv $5\!\! \times\!\! 5$, s2 & ConvTranspose $3\!\! \times\!\! 3$, s1 \\
        \hline
    \end{tabular}
    
    \caption{Details of $G_a$, $G_s$, $H_a$, $H_s$ modules in our lossless compression model.}
    \label{tab:lossless model}
    \vspace{-1em}
\end{table}

\section{Experiments}
\vspace{-0.5em}
\subsection{Setup}
\textbf{Datasets}
Our training dataset consists of $8000$ images chosen from ImageNet~\cite{deng2009imagenet} validation set, we test all methods on Kodak~\cite{kodak1993} dataset. During training, the RGB images are first randomly cropped into $256 \times 256$ size and then compressed into JPEG files with QP 75. During testing, the RGB images are directly compressed into JPEG files with QP 75.
\quad\\
\textbf{Implementations details}
We implement our experiments on PyTorch. We adopt a two-step training strategy. We first train our lossless compression model to converge on quantized DCT coefficients quantized by standard QP 75 quantization table. Then, for specific $\lambda_r, \lambda_d$ pair in equation \ref{eq:L_q}, we iteratively finetune the lossless compression model and learned quantization/inverse quantization tables which are initialized with standard QP 75 quantization table. We use Adam optimizer with $\beta_1 = 0.9$, $\beta_2 = 0.999$ and set batch-size to $16$ for training. In the first stage, we set the initial learning rate to $10^{-4}$, training epochs to $2000$, and decay the learning rate by $\gamma = 0.9$ after every $500$ epochs. In the second stage, we finetune the lossless compression model by setting the initial learning rate to $10^{-3}$, training epochs to $50$, and decaying the learning rate by $\gamma = 0.94$ after every epoch. For learned quantization/inverse quantization tables, we finetune it by setting the initial learning rate to $5e-3$, training epochs to $30$, and decaying the learning rate by $\gamma = 0.8$ after every epoch. We iteratively finetune the lossless compression model and the learned quantization/inverse quantization tables for total of $8$ times. By setting $(\lambda_r, \lambda_d) = \{(0.1, 100), (0.2, 100), (0.4, 100), (0.6, 100), (0.8, 100), (1, \\
100), (2, 100), (4, 100), (8, 100), (16, 100), (32, 100), (48, 50), \\
(64, 20), (64, 10), (64, 5)\}$, we obtain a series of models under different rate-distortion trade-offs.

\vspace{-0.5em}
\subsection{Performance}
We compare the proposed model against the state-of-art learned based image lossy compression method ELIC~\cite{he2022elic} and handcrafted image lossy compression method JPEG-XL~\cite{alakuijala2019jpeg}. As shown in Fig. \ref{fig:comparison}, our model can achieve arbitrarily high PSNR and MS-SSIM while the rate never exceeds the rate of the lossless compression model in the proposed framework. It means that the proposed model unifies the lossless compression and lossy compression for JPEG images. $D = 0$, namely PSNR equal to $+\infty$ or MS-SSIM equal to $1$, denotes the lossless compression while $D > 0$ denotes the lossy compression. For the proposed model, there is no longer a gap between lossless compression and lossy compression for JPEG images. In the high rate region, namely near the lossless compression rate, our model significantly outperforms ELIC~\cite{he2022elic} and JPEG-XL~\cite{alakuijala2019jpeg}. 

\vspace{-0.5em}
\subsection{Ablation study}
\begin{figure}
    \centering
    \includegraphics[scale=0.45]{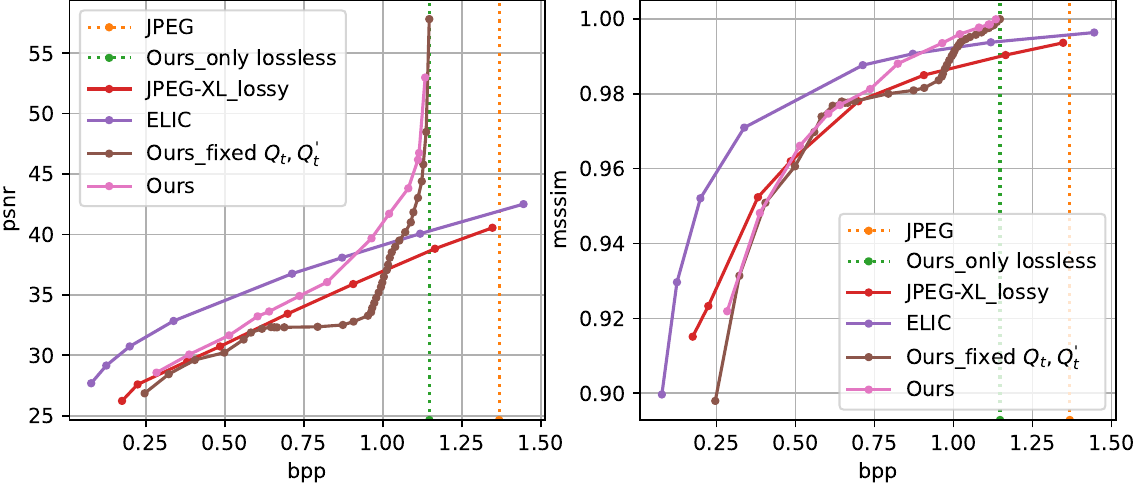}
    \caption{Objective quality comparison on Kodak dataset when recompressing JPEG images of QP 75.}
    \label{fig:comparison}
    \vspace{-1em}
\end{figure}

We test the proposed model with a fixed quantization table and inverse quantization table, which are set to the JPEG quantization table with QP from $10$ to $74$. As shown in Fig. \ref{fig:comparison}, the proposed model with learned $Q_t$, $Q_t^{'}$ significantly outperforms the one with fixed $Q_t$, $Q_t^{'}$, which verifies the effectiveness of learned quantization table and inverse quantization table for downstream lossless compression model. 

\section{Conclusion}
\vspace{-0.5em}
In this paper, we  reveal that there is a gap between lossless compression and lossy compression for JPEG recompression. To solve this problem, we propose an efficient JPEG recompression framework, which firstly unifies the lossless compression and lossy compression for JPEG images. Our framework consists of one learned quantization table, one learned inverse quantization table, and a learned lossless compression model in the DCT domain. The proposed model can achieve arbitrarily low distortion while the rate never exceeds the upper bound, namely the rate of the lossless compression model in the framework.

A major limitation of this study is that the rate-distortion performance falls too fast as the rate goes down. In the future, we will dive deeper into this issue and extend our method to other specialized data compression and transcoding problem.

\section{ACKNOWLEDGMENTS}
\vspace{-0.5em}
This work is supported by Baidu Inc. through Apollo-AIR Joint Research Center.

\vfill\pagebreak

% References should be produced using the bibtex program from suitable
% BiBTeX files (here: strings, refs, manuals). The IEEEbib.bst bibliography
% style file from IEEE produces unsorted bibliography list.
% -------------------------------------------------------------------------
\bibliographystyle{IEEEbib}
\bibliography{refs}

\begin{thebibliography}{10}

\bibitem{pennebaker1992jpeg}
William~B Pennebaker and Joan~L Mitchell,
\newblock {\em JPEG: Still image data compression standard},
\newblock Springer Science \& Business Media, 1992.

\bibitem{rabbani2002jpeg2000}
Majid Rabbani,
\newblock ``Book review: Jpeg2000: Image compression fundamentals, standards
  and practice,'' 2002.

\bibitem{boutell1997png}
Thomas Boutell,
\newblock ``Png (portable network graphics) specification version 1.0,''
\newblock Tech. {R}ep., 1997.

\bibitem{lian2012webp}
Li~Lian and Wei Shilei,
\newblock ``Webp: A new image compression format based on vp8 encoding,''
\newblock {\em Microcontrollers \& Embedded Systems}, vol. 3, 2012.

\bibitem{bellard2015bpg}
Fabrice Bellard,
\newblock ``Bpg image format,'' \url{https://bellard. org/bpg}, 2015,
\newblock Accessed: 2022-11-16.

\bibitem{ohm2018versatile}
Jens-Rainer Ohm and Gary~J Sullivan,
\newblock ``Versatile video coding--towards the next generation of video
  compression,''
\newblock in {\em Picture Coding Symposium}, 2018, vol. 2018.

\bibitem{Balle2017ICLR}
Johannes Ball{\'e}, Valero Laparra, and Eero~P. Simoncelli,
\newblock ``End-to-end optimized image compression,''
\newblock in {\em International Conference on Learning Representations}, 2017.

\bibitem{Balle2018ICLR}
Johannes Ballé, David Minnen, Saurabh Singh, Sung~Jin Hwang, and Nick
  Johnston,
\newblock ``Variational image compression with a scale hyperprior,''
\newblock in {\em International Conference on Learning Representations}, 2018.

\bibitem{Minnen2018NIPS}
David Minnen, Johannes Ball{\'e}, and George~D Toderici,
\newblock ``Joint autoregressive and hierarchical priors for learned image
  compression,''
\newblock {\em Advances in neural information processing systems}, vol. 31,
  2018.

\bibitem{Cheng2020CVPR}
Zhengxue Cheng, Heming Sun, Masaru Takeuchi, and Jiro Katto,
\newblock ``Learned image compression with discretized gaussian mixture
  likelihoods and attention modules,''
\newblock in {\em Proceedings of the IEEE/CVF Conference on Computer Vision and
  Pattern Recognition}, 2020, pp. 7939--7948.

\bibitem{Gao2021ICCV}
Ge~Gao, Pei You, Rong Pan, Shunyuan Han, Yuanyuan Zhang, Yuchao Dai, and Hojae
  Lee,
\newblock ``Neural image compression via attentional multi-scale back
  projection and frequency decomposition,''
\newblock in {\em Proceedings of the IEEE/CVF International Conference on
  Computer Vision}, 2021, pp. 14677--14686.

\bibitem{he2022elic}
Dailan He, Ziming Yang, Weikun Peng, Rui Ma, Hongwei Qin, and Yan Wang,
\newblock ``Elic: Efficient learned image compression with unevenly grouped
  space-channel contextual adaptive coding,''
\newblock in {\em Proceedings of the IEEE/CVF Conference on Computer Vision and
  Pattern Recognition}, 2022, pp. 5718--5727.

\bibitem{horn2017design}
Daniel~Reiter Horn, Ken Elkabany, Chris Lesniewski-Lass, and Keith Winstein,
\newblock ``The design, implementation, and deployment of a system to
  transparently compress hundreds of petabytes of image files for a
  file-storage service,''
\newblock in {\em 14th USENIX Symposium on Networked Systems Design and
  Implementation (NSDI 17)}, 2017, pp. 1--15.

\bibitem{alakuijala2019jpeg}
Jyrki Alakuijala, Ruud Van~Asseldonk, Sami Boukortt, Martin Bruse, Iulia-Maria
  Comșa, Moritz Firsching, Thomas Fischbacher, Evgenii Kliuchnikov, Sebastian
  Gomez, Robert Obryk, et~al.,
\newblock ``Jpeg xl next-generation image compression architecture and coding
  tools,''
\newblock in {\em Applications of Digital Image Processing XLII}. SPIE, 2019,
  vol. 11137, pp. 112--124.

\bibitem{guo2022practical}
Lina Guo, Xinjie Shi, Dailan He, Yuanyuan Wang, Rui Ma, Hongwei Qin, and Yan
  Wang,
\newblock ``Practical learned lossless jpeg recompression with multi-level
  cross-channel entropy model in the dct domain,''
\newblock in {\em Proceedings of the IEEE/CVF Conference on Computer Vision and
  Pattern Recognition}, 2022, pp. 5862--5871.

\bibitem{yeung2008information}
Raymond~W Yeung,
\newblock {\em Information theory and network coding},
\newblock Springer Science \& Business Media, 2008.

\bibitem{reed2017ms-pixelcnn}
Scott Reed, A{\"a}ron Oord, Nal Kalchbrenner, Sergio~G{\'o}mez Colmenarejo,
  Ziyu Wang, Yutian Chen, Dan Belov, and Nando Freitas,
\newblock ``Parallel multiscale autoregressive density estimation,''
\newblock in {\em International Conference on Machine Learning}. PMLR, 2017,
  pp. 2912--2921.

\bibitem{kingma2019bitswap}
Friso Kingma, Pieter Abbeel, and Jonathan Ho,
\newblock ``Bit-swap: Recursive bits-back coding for lossless compression with
  hierarchical latent variables,''
\newblock in {\em International Conference on Machine Learning}. PMLR, 2019,
  pp. 3408--3417.

\bibitem{mentzer2019l3c}
Fabian Mentzer, Eirikur Agustsson, Michael Tschannen, Radu Timofte, and Luc~Van
  Gool,
\newblock ``Practical full resolution learned lossless image compression,''
\newblock in {\em Proceedings of the IEEE/CVF conference on computer vision and
  pattern recognition}, 2019, pp. 10629--10638.

\bibitem{berg2021idf++}
Rianne van~den Berg, Alexey~A. Gritsenko, Mostafa Dehghani, Casper~Kaae
  S{\o}nderby, and Tim Salimans,
\newblock ``Idf++: Analyzing and improving integer discrete flows for lossless
  compression,''
\newblock in {\em International Conference on Learning Representations}, 2021.

\bibitem{deng2009imagenet}
Jia Deng, Wei Dong, Richard Socher, Li-Jia Li, Kai Li, and Li~Fei-Fei,
\newblock ``Imagenet: A large-scale hierarchical image database,''
\newblock in {\em 2009 IEEE conference on computer vision and pattern
  recognition}. Ieee, 2009, pp. 248--255.

\bibitem{kodak1993}
Eastman Kodak,
\newblock ``Kodak lossless true color image suite (photocd pcd0992),'' \url{},
  1993.

\end{thebibliography}

\end{document}